\documentclass[conference]{IEEEtran}
\IEEEoverridecommandlockouts
\usepackage{cite}
\usepackage{amsmath,amssymb,amsfonts}
\usepackage{algorithmic}
\usepackage{graphicx}
\usepackage{textcomp}
\usepackage{xcolor}
\usepackage{tabularx}
\usepackage{booktabs}
\usepackage{bbm}
\usepackage{url}
\def\BibTeX{{\rm B\kern-.05em{\sc i\kern-.025em b}\kern-.08em
    T\kern-.1667em\lower.7ex\hbox{E}\kern-.125emX}}

\makeatletter
\newcommand{\linebreakand}{%
  \end{@IEEEauthorhalign}
  \hfill\mbox{}\par
  \mbox{}\hfill\begin{@IEEEauthorhalign}
}
\makeatother

\begin{document}

\bstctlcite{IEEEexample:BSTcontrol}

\title{\huge SigmaScheduling: Uncertainty-Informed Scheduling of Decision Points for Intelligent Mobile Health Interventions\\
\thanks{*Equal contribution. This research was supported by the National Institutes of Health (K99EB037411, P50DA054039, P41EB028242, UH3DE028723, R01HL125440, P30AG073107, and 5P30AG073107-03 GY3 Pilots).} 
}

\author{Asim H.\ Gazi*, \textit{Member, IEEE}, Bhanu Teja Gullapalli*, \\
Daiqi Gao, Benjamin M.\ Marlin, Vivek Shetty, and Susan A.\ Murphy 

\thanks{A.\ H.\ Gazi (agazi@seas.harvard.edu), B.\ T.\ Gullapalli, D.\ Gao, and S.\ A.\ Murphy are with the Department of Statistics and the School of Engineering and Applied Sciences, Harvard University, Cambridge, MA, USA.   S.\ A.\ Murphy holds concurrent appointments at
Harvard University and as an Amazon Scholar. This
paper describes work performed at Harvard University
and is not associated with Amazon.
}
\thanks{B.\ M.\ Marlin is with the Manning College of Information and Computer Sciences, University of Massachusetts Amherst, Amherst, MA, USA.
    }
\thanks{V. Shetty is with the School of Dentistry, University of California, Los Angeles, Los Angeles, CA, USA.
    }
}

\maketitle

\begin{abstract}
Timely decision making is critical to the effectiveness of mobile health (mHealth) interventions. At predefined timepoints called ``decision points," intelligent mHealth systems such as just-in-time adaptive interventions (JITAIs) estimate an individual's biobehavioral context from sensor or survey data and determine whether and how to intervene. For interventions targeting habitual behavior (e.g., oral hygiene), effectiveness often hinges on delivering support shortly before the target behavior is likely to occur. Current practice schedules decision points at a fixed interval (e.g., one hour) before user-provided behavior times, and the fixed interval is kept the same for all individuals. However, this one-size-fits-all approach performs poorly for individuals with irregular routines, often scheduling decision points after the target behavior has already occurred, rendering interventions ineffective. In this paper, we propose SigmaScheduling, a method to dynamically schedule decision points based on uncertainty in predicted behavior times. When behavior timing is more predictable, SigmaScheduling schedules decision points closer to the predicted behavior time; when timing is less certain, SigmaScheduling schedules decision points earlier, increasing the likelihood of timely intervention. We evaluated SigmaScheduling using real-world data from 68 participants in a 10-week trial of Oralytics, a JITAI designed to improve daily toothbrushing. SigmaScheduling increased the likelihood that decision points preceded brushing events in at least 70\% of cases, preserving opportunities to intervene and impact behavior. Our results indicate that SigmaScheduling can advance precision mHealth, particularly for JITAIs targeting time-sensitive, habitual behaviors such as oral hygiene or dietary habits.

\end{abstract}


\begin{figure}
\includegraphics[width=240pt,keepaspectratio]{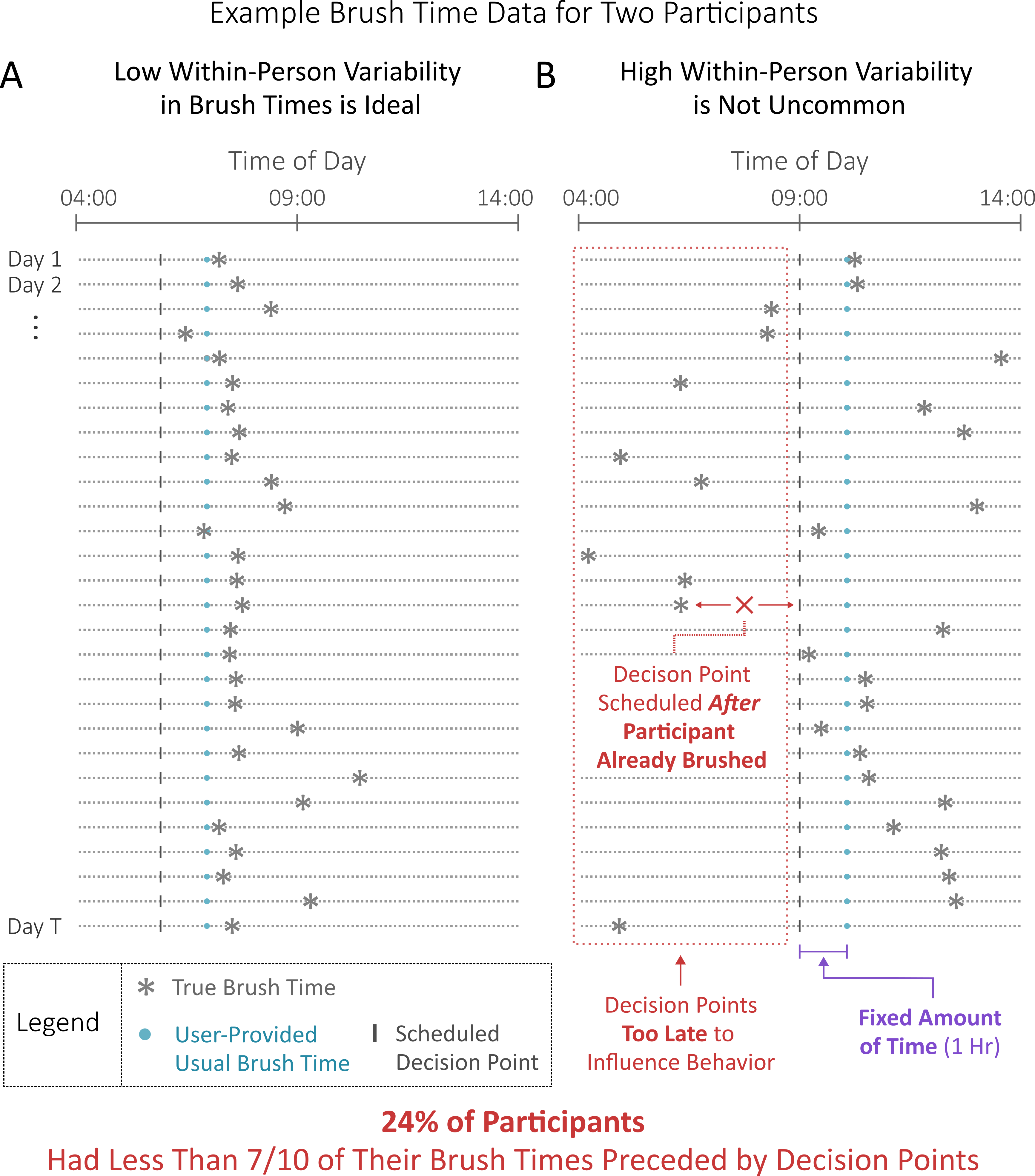}
\vspace{-5pt}
\caption{Example data illustrating heterogeneity in the predictability of routines for two participants in a trial of ``Oralytics." For each participant, 28 days of morning brush time data are shown. Days are stacked vertically, with each day's times plotted horizontally. (A) Data from a participant that exhibited less variability in brush times with respect to their user-provided times. (B) Data from a participant that exhibited more variability.}
\vspace{-15pt}
\label{intro_fig}
\end{figure}

\section{Introduction}
At predefined timepoints called ``decision points," just-in-time adaptive interventions (JITAIs) intelligently decide whether and how to provide mobile health support based on biobehavioral sensor or survey data \cite{Nahum-Shani2018}. A challenge for many JITAIs is that their effectiveness can depend on delivering interventions shortly before a target behavior, but identifying such pre-behavior time windows in real-time is often infeasible. For instance, while wrist-worn sensors can be used to detect eating \cite{Bell2020}, a reminder to take medication prior to a meal would be rendered ineffective if triggered only after the meal has already begun \cite{Ahmed2023}. More commonly, the challenge arises due to practical constraints in implementation. For the ``Oralytics" JITAI, which leverages a commercially available smart toothbrush and a companion app to promote oral self-care \cite{Trella2025}, data from the brush becomes available only once each night. As a result, decision points for a given day must be scheduled in advance based on prior brushing patterns to anticipate -- rather than respond to -- forthcoming behavior.

\begin{figure*}
\includegraphics[width=\textwidth,keepaspectratio]{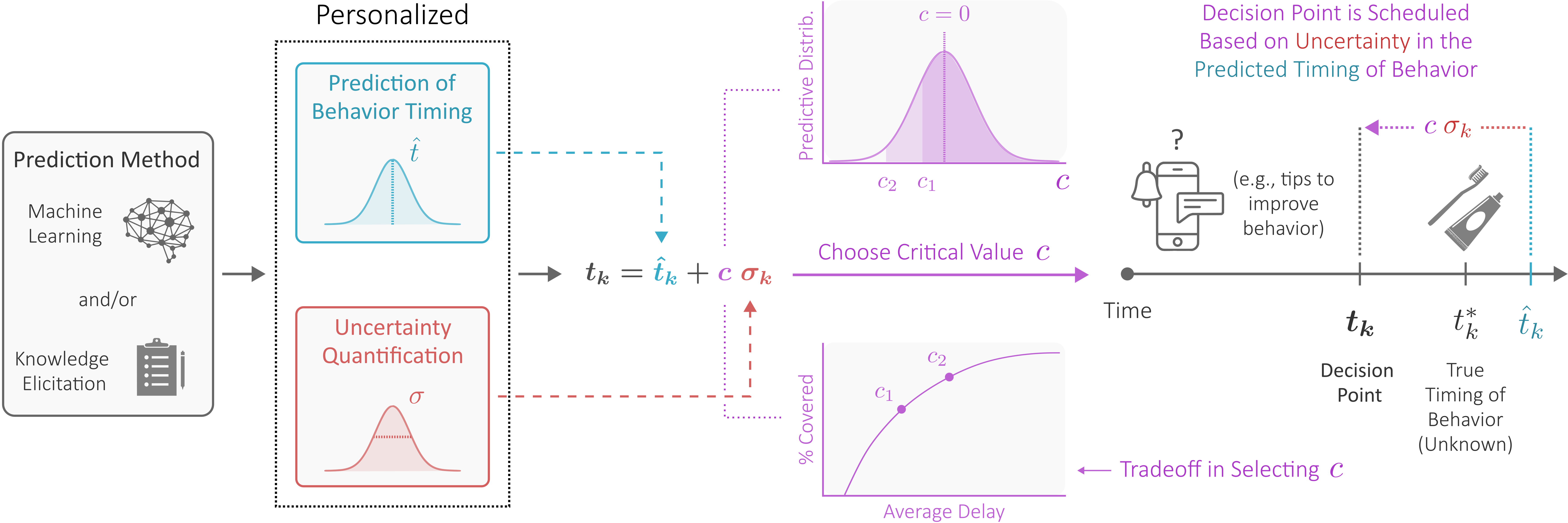}
\vspace{-16pt}
\caption{SigmaScheduling approach. SigmaScheduling uses personalized predictions of behavior times -- along with quantifications of uncertainty -- to schedule decision points according to the probability of preceding the target behavior. The critical value, c, dictates the number of uncertainty measures (e.g., standard deviations) away from the mean that decision points are scheduled. Choosing c involves a tradeoff between the proportion of true behavior times preceded or ``covered" by decision points and the average delay from decision points to behavior times. Based on the choice of c, SigmaScheduling takes predicted behavior times and the uncertainty quantified at each timestep, k, and schedules the kth decision point accordingly.}
\label{methods_fig}
\vspace{-8pt}
\end{figure*}

The status quo approach to scheduling decision points is to ask individuals to provide expected behavior times prior to JITAI deployment and then schedule decision points the same fixed duration -- for all users -- before expected behavior times \cite{Nahum-Shani2024, Zhao2023, Fuchs2025}. While this status quo may perform satisfactorily for individuals with relatively predictable routines, the status quo often falters for more variable behavior patterns, as shown in Fig. \ref{intro_fig}. Increasing the fixed duration from one to four hours could address the issue in Fig. \ref{intro_fig}B -- but would unnecessarily increase the delay from decision points to brush times for more predictable routines (e.g., Fig. \ref{intro_fig}A). In settings where behavior predictability varies widely across the population, the one-size-fits-all status quo can squander valuable intervention opportunities. For example, \textit{nearly a quarter of participants in the first deployment of Oralytics had 70\% or fewer of their brushing events preceded by a decision point.}

This paper proposes and evaluates SigmaScheduling: a method to dynamically schedule decision points based on uncertainty in predicted behavior times. Our hypothesis is that SigmaScheduling will more effectively precede target behaviors with decision points by scheduling decision points earlier when routines are less predictable and closer beforehand otherwise. To test this hypothesis, we apply SigmaScheduling to real-world data from 68 participants of a 10-week trial of Oralytics \cite{Trella2025}, using (1) user-provided times and (2) online Bayesian machine learning (ML) for brush time prediction and uncertainty quantification (UQ). Our contribution is to demonstrate that for both approaches, SigmaScheduling increases the proportion of participants whose brushing events are preceded by a decision point at least 70\% of the time.

\section{Methods}

\subsection{SigmaScheduling}
\label{subsec_sigmaSched}
Fig. \ref{methods_fig} illustrates SigmaScheduling. At each opportunity to schedule a decision point targeting a predicted behavior, indexed by $k \in \{1, 2, ..., K\}$, SigmaScheduling needs a prediction of the behavior time, $\hat{t}_k$, and UQ with an uncertainty measure, $\sigma_k$. The decision point, $t_k$, is scheduled via (\ref{eq_SigmaScheduling}).
\begin{equation}
    t_k = \hat{t}_k + c \,\sigma_k
    \label{eq_SigmaScheduling}
\end{equation}
The true behavior time, $t^*_k$, is forecasted via $\hat{t}_k$, and $c$ is a design parameter. The status quo approach instead schedules via $t_k = \hat{t}_k + F$, where $F \leq 0$ is a fixed duration. Effectively, SigmaScheduling replaces $1$ multiplied by a constant (i.e., $\hat{t}_k + 1*F$) with $\sigma_k$ multiplied by a constant (i.e., $\hat{t}_k + \sigma_k * c$). 

The key design choice for SigmaScheduling is the parameter $c$. Theoretically, $c$ represents a \textit{critical value} and can be selected based on the desired probability that $t_k^* > t_k$. For example, if the posterior distribution $P(t_k^*|\hat{t}_k)$ is modeled as a Gaussian distribution, then $c$ represents a z-score. In practice, selecting $c$ involves a tradeoff between ``coverage," the proportion of behavior times preceded by decision points (i.e., $\frac{1}{K} \cdot P$, where $P = \sum_{k=1}^{K}{\mathbbm{I} [t_k < t^*_k] }$), and the mean delay from decision points to behavior times (i.e., $\frac{1}{P} \sum_{k = 1}^{K}{\mathbbm{I} [t_k < t^*_k] \cdot (t^*_k - t_k)}$). In general, $c$ can vary across time (i.e., $\{c_k\}_{k=1}^K$) or individuals, but in this work, $c$ is fixed.

\begin{figure*}
\includegraphics[width=\textwidth,keepaspectratio]{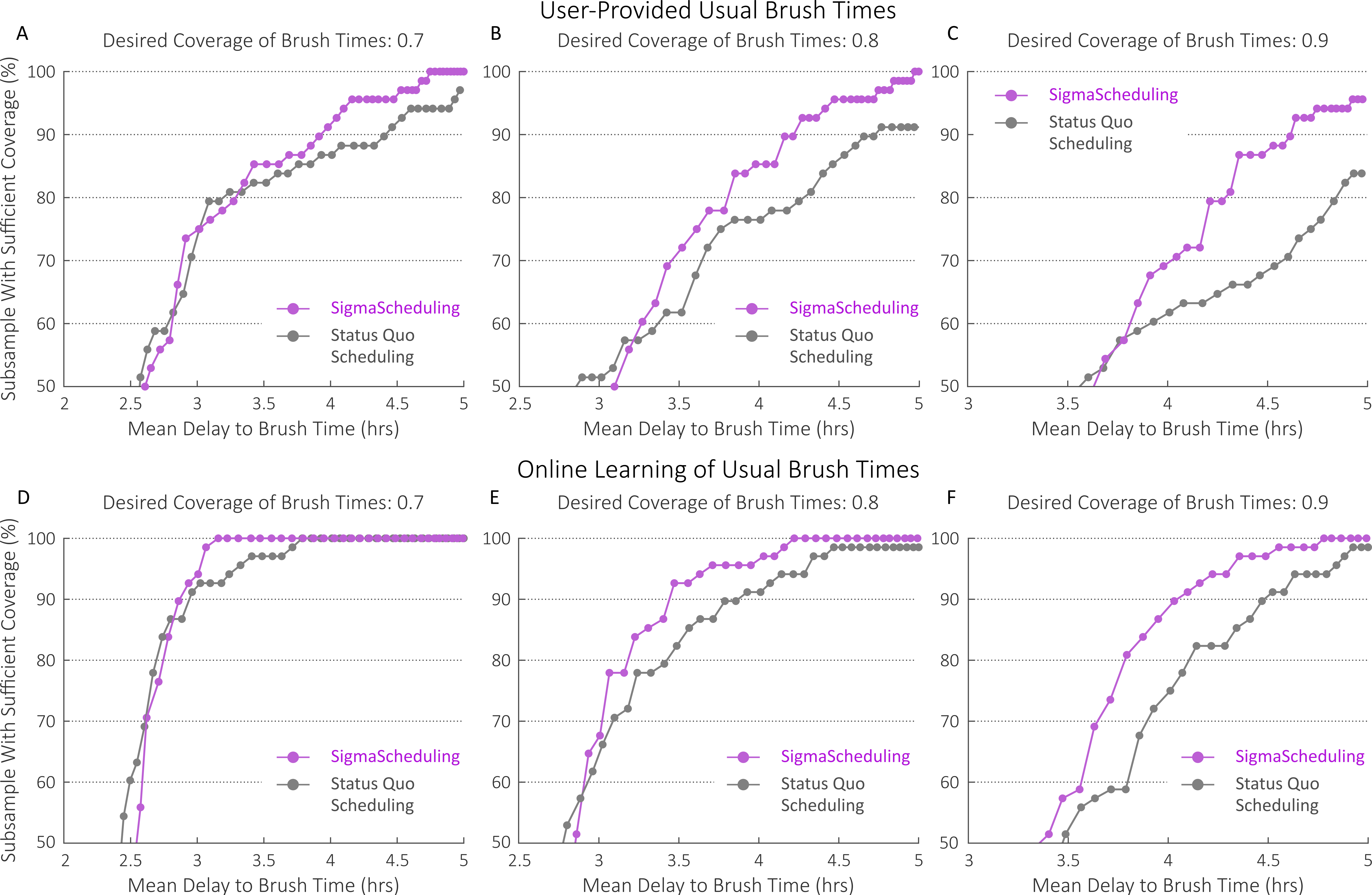}
\vspace{-17pt}
\caption{Comparison of SigmaScheduling with Status Quo Scheduling for select coverage cutoffs. Note that the x-axes differ between columns of the subplot grid (i.e., A and D have the same x-axis, but B does not). The delays quantified along the x-axis refer to means across all participants of each participant-specific mean delay from decision points to the brush times that were successfully covered. Subsamples shown on the y-axis refer to the proportion of 68 participants that met the desired coverage of brush times specified for each subplot. For example, a y-value of 89.7\% indicates that 61 participants' decision points were scheduled to satisfy the desired coverage.  Each point on the SigmaScheduling curves corresponds to a different critical value, $c$, used in decision point scheduling, while each point on the status quo scheduling curves represents a possible fixed duration, $F$. Curves illustrate the tradeoff between coverage and delay and compare the two methods across $c$ and $F$. Similar to receiver operating characteristic (ROC) curves, the optimal curve would be a horizontal line at a y-value of 100\%. (A) Comparison between scheduling approaches when using user-provided usual brush times with a desired coverage of brush times per participant of 0.7. (B) Comparison using user-provided usual brush times with a desired coverage of 0.8. (C) Comparison using user-provided usual brush times with a desired coverage of 0.9. (D) Comparison using machine learning (ML) model predictions of usual brush times with a desired coverage of 0.7 for each participant. (E) Comparison using ML with a desired coverage of 0.8. (F) Comparison using ML with a desired coverage of 0.9.}
\label{results_fig}
\vspace{-10pt}
\end{figure*}

\subsection{Prediction and UQ with User-Provided Times}
\label{subsec_UQuserProv}
User-provided times are one avenue of predicting behavior times. A simple approach for UQ is to compute $\sigma_k$ via the standard error for a prediction interval based on past errors in user-provided times \cite{Geisser1993}. In particular, $\sigma_k$ in (\ref{eq_SigmaScheduling}) can be computed via $\sigma_k = s_\epsilon \sqrt{1 + 1/k}$, where $s_\epsilon = \sqrt{\frac{1}{k - 1} \sum_{i=1}^{k} (\epsilon_i - \bar{\epsilon}_k)^2}$ is the standard deviation (SD) of past errors accumulated until timestep $k$, with mean $\bar{\epsilon}_k = \frac{1}{k}\sum_{i=1}^{k}{\epsilon_i}, \: \epsilon_i = t^*_i - \hat{t}_i$. In this work, we set $\sigma_1 = 0$. The approach described forms a (biased) prediction interval based on the sample SD of residuals, which likens $c$ in (\ref{eq_SigmaScheduling}) to critical values of the t-distribution. 

\subsection{Prediction and UQ via Online Bayesian ML}
\label{subsec_BayesianML}
Another avenue of predicting behavior times (i.e., $\hat{t}_k$) is via ML. In this work, we focus on online Bayesian ML to match the approach planned for the next iteration of Oralytics. Online learning enables a trained ML model to personalize to the target individual as more of the individual's data is observed. Bayesian algorithms produce posterior predictive distributions, $P(t^*_k|\hat{t}_k)$, from which the SD can be used for $\sigma_k$.

\subsection{Oralytics Data Used for Evaluation}
\label{subsec_Oralytics}
This paper evaluates SigmaScheduling using brush time data from 68 participants in the first trial of Oralytics \cite{Trella2025}. Of the 72 participants analyzed by Trella et al., we excluded one with less than a week of analyzable brush times and three with corrupted data due to user-provided brush times between 0:00 and 04:00. Participants were adults at risk for dental disease. The 70-day micro-randomized trial had two daily decision points (morning and evening); was approved by the University of California, Los Angeles Institutional Review Board (IRB\#21–001471) and registered on ClinicalTrials.gov (NCT05624489) \cite{Nahum-Shani2024}; and lasted Sept.\ 2023 to July 2024 \cite{Trella2025}.


Participants provided typical morning and evening brush times on weekdays and weekends. Ground truth brush times, $\{t^*_k\}_{k=1}^K$, were sensed via smartbrush (Oral-B 8000). Only brush times, $t^*_k$, without prior intervention for the same $k$ were analyzed to focus on self-determined behavior times, which decision points should be scheduled to precede. A mean of 53 $\pm$ 23 ($\pm$ SD) brush times were analyzed per participant.

\subsection{Online Bayesian Linear Regression for Oralytics}
\label{subsec_BLR}
\vspace{-2pt}
Leave-one-participant-out cross validation was used to produce $N$ participant-specific Bayesian linear regression (BLR) models for brush time prediction and UQ. BLR was used to match the upcoming deployment of Oralytics. Gaussian priors for regression weights were set as in prior work \cite{Trella2025}. Prior means were set via a generalized estimating equation (GEE) fit to the $N-1$ participants' training data: the GEE point estimate was used for statistically significant regressors; zero otherwise. Prior SDs were set by fitting a GEE model to each training participant's data: the SD of point estimates was used for statistically significant regressors; half the SD otherwise. An inverse-gamma prior was used for the observation noise variance. The regressors were user-provided brush times; day of the week; morning or evening; the minimum, maximum, and coefficient of variation of past week brush times, most recent brush time; and number of no-brush days. For the held-out participant, priors were updated online with $\{t^*_i\}_{i=1}^{k-1}$ before predicting $\hat{t}_k$ and producing Gaussian posterior, $P(t^*_k|\hat{t}_k)$.

\subsection{Comparing SigmaScheduling Versus Status Quo}
\label{subsec_comparison}
We compared SigmaScheduling and status quo scheduling using tradeoff curves analogous to receiver operating characteristic (ROC) curves. The ROC-like curves plotted the proportion of participants with the desired coverage of brush times versus the mean delay across participants from decision points to brush times, varying $c$ from 0 to –3 for SigmaScheduling and $F$ from 0 to –6 hours for status quo. To align with Oralytics constraints, decision points scheduled before 04:00 for morning brushing and before 16:00 for evening brushing were clipped. Scheduling performance is improved if the proportion of participants with desired coverage increases for a given mean delay, analogous to increasing the true positive rate for a given false positive rate on a ROC curve. Similarly, the areas under the the ROC-like curves can be compared, analogous to comparing ML classifiers using areas under ROC curves. Areas were bounded below by 50\% (i.e., at least a majority achieve desired coverage) and to the right by 5 hours.

The threshold for desired coverage per participant was varied from 0.66 to 0.99. The lower limit of 0.66 aligns with the threshold of 2/3 used for feasibility analyses \cite{Coughlin2024}. We evaluated the percentage of participants with desired coverage, rather than the mean coverage, because we aim to achieve the desired coverage for a majority of participants, rather than to sacrifice coverage for some participants in exchange for others (e.g., 0.7 coverage for two participants versus 0.9 and 0.5).

\begin{table}
\caption{Area Under ROC-Like Curves as Function of Desired Coverage}
\label{results_table}
\centering
\vspace{-6pt}
\setlength{\tabcolsep}{4pt}
\begin{tabularx}{\linewidth}{l>{\centering\arraybackslash}X>{\centering\arraybackslash}X|>{\centering\arraybackslash}X>{\centering\arraybackslash}X}
\toprule
& \multicolumn{2}{c|}{User-Provided} & \multicolumn{2}{c}{Model-Predicted} \\
Desired & \textit{Status} & \textit{$\Sigma$-} & \textit{Status} & \textit{$\Sigma$-} \\
Coverage & \textit{Quo} & \textit{Scheduling} & \textit{Quo} & \textit{Scheduling} \\
\midrule
0.66 & 86.13 & \textbf{92.77} & \textbf{121.18} & 119.74 \\
0.70 & 77.94 & \textbf{85.51} & 112.21 & \textbf{113.07} \\
0.75 & 64.61 & \textbf{75.24} & 95.31 & \textbf{100.93} \\
0.80 & 49.15 & \textbf{62.94} & 80.38 & \textbf{89.38} \\
0.85 & 38.53 & \textbf{54.00} & 66.00 & \textbf{77.67} \\
0.90 & 22.71 & \textbf{39.67} & 45.17 & \textbf{59.48} \\
0.95 & 6.16 & \textbf{18.17} & 15.44 & \textbf{34.84} \\
0.99 & 0.00 & 0.00 & 0.02 & \textbf{3.84} \\
\bottomrule
\vspace{-7pt} \\
\multicolumn{5}{l}{\textbf{Boldface} indicates superior performance. Rows are not comparable.} \\
\vspace{-17pt}
\end{tabularx}
\end{table}

\section{Results}
Fig. \ref{results_fig} compares status quo scheduling and SigmaScheduling via ROC-like curves described in \ref{subsec_comparison}. Interpreting Fig. \ref{results_fig}B as an example, if the goal is for participants to have at least 8/10 of their brush times preceded by decision points, status quo scheduling achieves this at a mean delay of $\sim$4.5 hours for $\sim$85\% of participants. For the same mean delay, SigmaScheduling achieves the desired coverage for $\sim$95\% of participants. Table \ref{results_table} details the areas under the curves.

\section{Discussion}
Using 68 participants' real-world data from the first deployment of the Oralytics JITAI, we find that SigmaScheduling improves the likelihood that interventions are delivered at opportune moments, before the target behavior of tooth brushing, in over 7/10 of instances. The improvement stems from SigmaScheduling's core principle of personalization: decision points are scheduled a longer duration before predicted brush times for participants with less predictable routines, while using shorter durations otherwise. In comparison, status quo scheduling uses the same duration for all participants.


SigmaScheduling's advantage becomes more pronounced as desired coverage increases. However, under more lenient criteria such as when covering only 2/3 of brushing events per participant, status quo scheduling performs comparably or better. One of the primary reasons for this is that for participants with predictable routines, the majority of true brush times fall fairly close to predicted brush times. In this dataset, participants with irregular routines were the minority. If it is acceptable to exclude the minority of participants with less predictable routines, (i.e., the threshold for desired coverage is lower), status quo scheduling suffices.

This work has limitations. Our evaluation of SigmaScheduling is limited to retrospective analysis within the oral self-care domain. Future studies are needed to deploy SigmaScheduling, extend its application to domains such as dietary behavior \cite{Zhao2023, Fuchs2025}, and evaluate the behavioral impact of improved coverage. 
\section{Conclusion}
This paper introduces SigmaScheduling, a novel method for scheduling JITAI decision points that dynamically adapts the lead time between decision points and predicted behavior times to the uncertainty in predicted times. Applied to Oralytics data, SigmaScheduling more consistently schedules decision points prior to toothbrushing (i.e., during windows of receptivity \cite{Nahum-Shani2018}). Future work will evaluate SigmaScheduling in the upcoming clinical trial deployment of Oralytics and apply SigmaScheduling to other domains such as dietary behavior.

\bibliographystyle{ieeetr}
\bibliography{references_Gazi}

\end{document}